\documentclass[%
 reprint,
 amsmath,amssymb,
 aps,
]{revtex4-1}

\usepackage{caption}
\usepackage{subcaption}
\usepackage{color,soul}
\usepackage{graphicx}
\usepackage{dcolumn}
\usepackage{bm}

\begin{document}


\title{Space-Filling Curves as a Novel Crystal Structure Representation for Machine Learning Models}

\author{Dipti Jasrasaria}
\author{Edward O. Pyzer-Knapp}
\author{Dmitrij Rappoport}
\author{Al\'{a}n Aspuru-Guzik}
\email{aspuru@chemistry.harvard.edu}
\affiliation{Department of Chemistry and Chemical Biology \\
Harvard University, Cambridge, MA 02138, USA}

\begin{abstract}
A fundamental problem in applying machine learning techniques for chemical problems is to find suitable representations for molecular and crystal structures. While the structure representations based on atom connectivities are prevalent for molecules, two-dimensional descriptors are not suitable for describing molecular crystals. In this work, we introduce the SFC-M family of feature representations, which are based on Morton space-filling curves, as an alternative means of representing crystal structures. Latent Semantic Indexing (LSI) was employed in a novel setting to reduce sparsity of feature representations. The quality of the SFC-M representations were assessed by using them in combination with artificial neural networks to predict Density Functional Theory (DFT) single point, Ewald summed, lattice, and many-body dispersion energies of 839 organic molecular crystal unit cells from the Cambridge Structural Database that consist of the elements C, H, N, and O. Promising initial results suggest that the SFC-M representations merit further exploration to improve its ability to predict solid-state properties of organic crystal structures.
\end{abstract}

\pacs{Valid PACS appear here}

\maketitle

\section{\label{sec:level1}Introduction}

High-throughput virtual screening methods have become popular and have proven to be successful in the systematic and accelerated discovery of new materials\cite{AAG_2011,EOPK_HTVS,flow_battery,OLEDs,Halls_2010,Kanal,OBoyle,Shu,Colon,Halls_2013,Halls_2013v2,Kadantsev,Curtarolo,Korth,Wilmer,Jain,Gregor}. One prominent example is the Harvard Clean Energy Project (CEP) which is am \textit{in silico} high-throughput screening program for small molecule materials aimed at discovering novel organic photovoltaic materials (OPVs)\cite{AAG_2011}. The CEP evaluates the power conversion efficiency of candidates based on frontier molecular orbital (FMO) calculations\cite{Hachmann_2014}. These calculations are performed on donor molecules using quantum mechanical methods, such as Density Functional Theory (DFT). However, employing such methods on periodic systems such as crystal structures is substantially more computationally expensive. Therefore, such solid-state calculations can only be performed on a small number of candidates of interest.

Machine learning (ML) methods are a powerful tool for decreasing this computational burden by directly predicting the solid-state properties of candidates by training a model with representative data. The quality of ML methods depends on two major factors: the learning algorithm and the feature representation, which contains the descriptive attributes of the input data.

The past several years have seen significant advances in ML methods using two-dimensional molecular descriptors\cite{Geppert_SVM,Ouvrard}. However, quantitative structure-property relationships that rely on two-dimensional molecular descriptors may fail to predict properties that depend on three-dimensional structure (\textit{i.e.} molecular conformations or crystal structure polymorphs)\cite{Palmer,Mitchell}, which are important for the design of new materials, such as OPVs.

To address these shortcomings, ML methods using crystal structure descriptors have been developed over the past five years\cite{Schutt,Faber,Carr,Huan,Pilania} to predict solid-state properties. Many of these feature representations are versions of radial distribution functions\cite{Schutt}, atomic coordinates paired with lattice vectors\cite{Schutt}, Coulomb matrices\cite{Faber,hase_2016} or other graphical representations\cite{Carr,Huan}. These studies have demonstrated predictions with relatively high accuracies while requiring relatively small training sets and drastically decreasing computation times. However, these representations have been primarily developed and used for inorganic or hybrid inorganic-organic crystals. In contrast, space-filling curves provide a well-defined and invertible mapping between the physical three-dimensional space (or, more generally, a $N$-dimensional space) and the one-dimensional feature representation vector that is suitable for both periodic systems and isolated molecules. We believe that this is the first study that uses machine learning models to predict solid state properties of organic molecular crystals.

Space-filling curves, such as the Peano, Hilbert, and Morton curves, have found widespread use in diverse application areas, such as database access, parallel algorithms, geographic information systems, and image processing\cite{sagan, gaede, bader}. The key property of space-filling curves is that they ``flatten'' the multidimensional space while preserving locality (\textit{i.e.}, points close to each other in the multidimensional space are mapped onto close locations along the one-dimensional vector). As a result, they provide a lossless and unbiased encoding of the crystal structure. This is consistent with the data-driven paradigm\cite{Simonyan14c, lusci, Duvenaud} that is increasingly used in state-of-the-art machine learning applications, rather than the previously popular hand-crafted paradigm.

In this work, we demonstrate the usefulness of the \textbf{SFC-M} family of representations for machine learning applications by using it as an input to artificial neural networks in order to predict DFT-calculated energies on a dataset of 839 molecular unit crystals.

\section{\label{sec:level2}Dataset}

Using Conquest, the entire Cambridge Structural Database (CSD) \cite{csd} was searched for organic molecular crystals that would be simple, diverse, and have accurately determined structures. This was established by the following search constraints:
\begin{itemize}
    \item contain only elements C, N, H, and O
    \item R factor $\leq$ 5
    \item determination temperature $\leq$ 100K
    \item Z $\leq$ 4 (number of molecules in unit cell)
    \item Z' = 1 (number of distinct molecules in unit cell)
    \item exclude disorder, errors, polymers, ions
    \item all 3D coordinates are determined
\end{itemize}

This search yielded 939 crystal structure unit cells, for which Crystallographic Information Files (CIFs) were extracted. Crystal structures with charged atoms were identified using the canonical SMILES strings\cite{weininger_smiles_1988}, which were generated using Open Babel\cite{openb}, and removed. This left a set of 839 unit cells.

Quantum Espresso\cite{quantume} was used to calculate the single point energy and many-body dispersion (MBD) energy \cite{mbd1, mbd2} for each unit cell with the PBE\cite{pbe1, pbe2} functional, planewave basis set with energy cutoff of 400 Ry, and norm-conserving HSCV\cite{hscv1, hscv2, hscv3} pseudopotentials.

Multipoles up to the hexadecapole level were calculated for molecules in each unit cell at a B3LYP\cite{Becke_1988, Lee, Becke_1993}, 6-31G**\cite{hehre, hariharan} level of theory using Gaussian09\cite{g09} and GDMA\cite{gdma}. The Williams99 force field\cite{williams_improved_2001, w992, w993} with multipole electrostatics as implemented in DMACRYS\cite{dma_neigh} was used to perform a force-field based single point energy calculation on the unit cell, from which the Ewald summed energy and lattice energy was extracted.

These properties of crystals are important in materials discovery and to the pharmaceutical community because of their use in solubility prediction\cite{sol1, sol2, sol3} and crystal structure prediction\cite{csp, csp_blind1, csp_blind2, csp_blind3}.

\section{\label{sec:level3}Feature Representation}

The predictive ability of machine learning (ML) is largely dependent on two factors: the algorithm and the feature representations the algorithm receives as inputs.

Good feature representations are:
\begin{itemize}
    \item non-degenerate (\textit{i.e.}, two unique inputs should yield two distinct feature representations. This property is automatically satisfied if the mapping between the inputs and the representation is invertible.),
    \item invariant to trivial transformations (\textit{i.e.}, insensitive to translations and rotations of the system as a whole).
\end{itemize}

In the following, we describe the \textbf{SFC-M} family of feature representations, which is consistent with both of these properties.

\subsection{\label{sec:level4}Morton Space-Filling Curves}

The \textbf{SFC-M} family of representations maps a set of discretized $N$-dimensional coordinates to a one-dimensional vector while preserving locality\cite{morton}. We chose a four-dimensional structure representation, which consists of one atomic descriptor dimension and three discretized spatial dimensions. 
\begin{figure}[h!]
\includegraphics[width=\columnwidth]{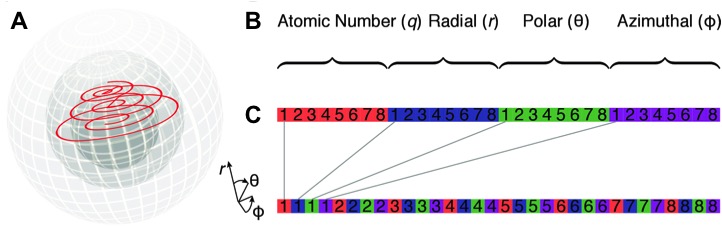}
\caption[Morton Space-Filling Curve]{The Morton Space-Filling Curve maps the four-dimensional coordinates that describe a crystal structure to a one-dimensional vector. \textbf{A}: The spatial coordinates are converted from Cartesian to spherical. \textbf{B}: The coordinates are discretized to a 4-D grid with an 8-bit resolution. \textbf{C}: The coordinates are flattened to a 1-D vector.
\label{fig:morton}}
\end{figure}

To generate the \textbf{SFC-M} representations, the crystal structure is first oriented along its principal moments of inertia and then the positions of its atoms are converted from Cartesian coordinates to spherical coordinates (A). Next, the dimensions are discretized to a four-dimensional grid with a 4-bit resolution (B, which shows a grid with an 8-bit resolution). Finally, coordinates on the grid are flattened to a one-dimensional vector (C). To flatten, coordinates are mapped to bins on a four-dimensional grid. Next, the bits for each bin are interleaved. The bins are flattened by traversing these interleaved values in ascending order, producing a recursive Z shape\cite{morton, orenstein}. When a bin contains the coordinates of an atom in the crystal structure, the corresponding index of the vector is incremented. The prototype Python code for performing structure encoding using the Morton space-filling curve is provided online\cite{molz}.

Our \textbf{SFC-M} family of representations considers a sphere of 60 \r{A} that is generated from the center of the unit cell of the crystal structure.

The atomic descriptor dimension of the crystal structure can be defined in numerous ways. This study explores three such definitions that yield three different versions of the \textbf{SFC-M} representation:
\begin{itemize}
    \item \textbf{SFC-M1}: atomic descriptor is the atomic number (\textit{i.e.}, an oxygen is assigned the number 8),
    \item \textbf{SFC-M2}: atomic descriptor is the atomic number plus the number of coordination sites (\textit{i.e.}, an oxygen bonded to two other atoms is assigned the number 12),
    \item \textbf{SFC-M3}: atomic descriptor is the Coulombic charge, $C_{\alpha}$, of an atom $\alpha$ given by
    \begin{equation}
        C_{\alpha} = \sum_{i=1}^{N_{\beta}}\frac{Z_{\alpha}Z_{\beta_{i}}}{r_{\alpha\beta_{i}}}
    \end{equation}
    where $N_{\beta}$ is the number of all other atoms in the unit cell, $Z_{x}$ is the nuclear charge of atom $x$, and $r_{xy}$ is the distance between atoms $x$ and $y$.
\end{itemize}

The flexibility of the atomic descriptor dimension is one of the strengths of the \textbf{SFC-M} representation because it allows for the injection of domain knowledge into the representation, potentially leading to more predictive machine learning.

\subsection{\label{sec:level4}Dimensionality Reduction}

Our \textbf{SFC-M} representations yield sparse (zero-filled) vectors of length $2^{16}$. Sparse vectors can hinder the ability of ML algorithms to identify patterns in data and make meaningful predictions. Latent Semantic Indexing (LSI)\cite{landauer_introduction_1998, laundauer2011} is a leading approach for transforming sparse vectors into smaller, denser representations. The \textbf{SFC-M} representations were treated using the ``bag of words" methodology, where each sparse vector (\textit{i.e.}, crystal structure) represents a word dictionary and each non-zero index (\textit{i.e.}, atom) represents a word that is present. The ``bag of words" model disregards the order in which words appear and only considers the frequency with which they appear. LSI maps the set of \textbf{SFC-M} representations to a lower-dimensional space that represents words which are commonly used (\textit{i.e.}, structural motifs in the set).

To do this, a counts matrix, $A$, which counts the number of instances for which an atom appears in each crystal structure, is generated. Next, the counts matrix is modified with Term Frequency--Inverse Document Frequency (TF--IDF) to yield a modified counts matrix, $A^{\prime}$, where the $i, j$th index is
\begin{equation}
    A^{\prime}_{i, j} = \frac{A_{i, j}}{\sum_{i}A_{i, j}} \log{\frac{D}{D_{i}}}
\end{equation}
where $D$ is the number of crystal structures in the set, and $D_{i}$ is the number of crystal structures in which atom $i$ appears. TF--IDF assigns weights based on atom frequency.

Finally, the $A^{\prime}$ matrix is decomposed using Singular Value Decomposition (SVD).

\begin{figure}[h!]
\includegraphics[width=\columnwidth]{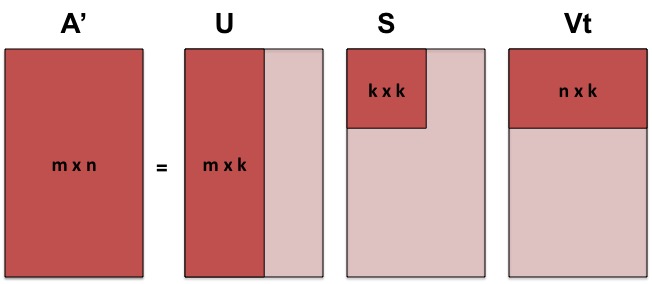}
\caption[Singular Value Decomposition]{Singular Value Decomposition decomposes the transformed counts matrix, $A'$, into three matrices, $U$, $S$, and $Vt$. The set of reduced \textbf{SFC-M} representations is equal to $S \cdot Vt$.
\label{fig:svd}}
\end{figure}

As illustrated in Fig. \ref{fig:svd}, the $A^{\prime}$ matrix represents a mapping of $n$ crystal structures on the $m$-dimensional atom space. This matrix is decomposed into the following matrices:
\begin{itemize}
    \item \textbf{$U$ matrix}: represents a mapping of the $m$ atoms on the reduced, $k$-dimensional space,
    \item \textbf{$S$ matrix}: a diagonal matrix that contains the singular values of the reduced space; the $k$th singular value describes the contribution of the $k$th dimension in the reduced space to the overall space,
    \item \textbf{$Vt$ matrix}: represents a mapping of the $n$ crystal structures on the reduced, $k$-dimensional space.
\end{itemize}

\begin{figure}[h!]
\includegraphics[width=\columnwidth]{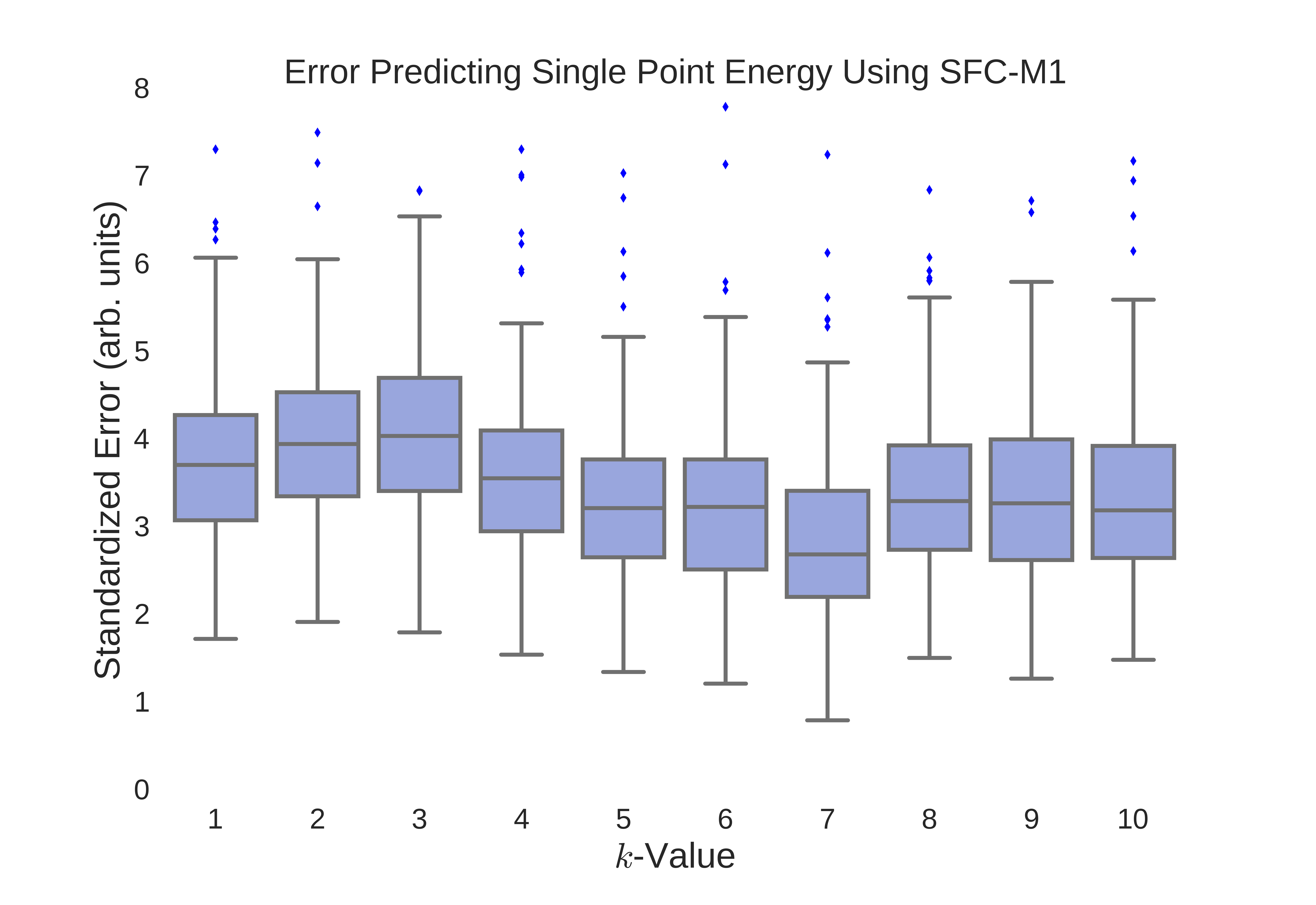}
\caption[Optimal $k$-value]{10-fold cross validated error in predicting single point energy using \textbf{SFC-M1} reduced using different $k$-values. All predictions were made using artificial neural networks of the same architecture.  This error is unitless.
\label{fig:k_vals}}
\end{figure}

One challenge presented by this method is determining a suitable $k$-value (number of reduced dimensions). As illustrated in Fig. \ref{fig:k_vals}, the $k$-value should be large enough that important motifs in the sparse vectors are retained. However, the $k$-value should be small enough that additional noise is not incorporated.

\begin{figure}[h!]
\includegraphics[width=\columnwidth]{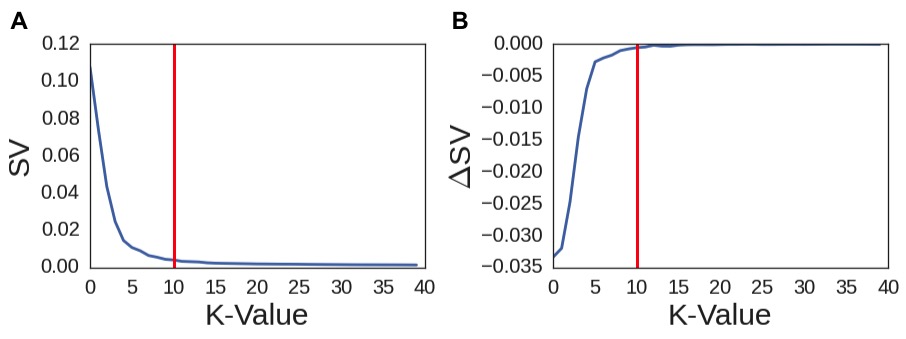}
\caption[Singular values]{The singular values (\textbf{A}) and gradient of the singular values (\textbf{B}) of \textbf{SFC-M2}. The chosen $k$-value, which corresponds to the vertical red line, represents a trade-off between accuracy and smoothness.
\label{fig:optimk}}
\end{figure}

The singular values (the diagonal of the $S$ matrix) can be used to choose a suitable $k$-value. Fig. \ref{fig:optimk} shows the plots of the singular values (A) and the gradient of the singular values (B) of \textbf{SFC-M2}.

The chosen $k$-value corresponds to the number of dimensions at which the gradient of the singular value curve is less than -0.001. For \textbf{SFC-M2}, we chose $k = $ 10 dimensions.

This implementation of LSI\cite{maml} takes advantage of the sparseness of the \textbf{SFC-M} representations so that computation time and memory scale with the number of non-zero indices, not the total length of the sparse vectors.

\section{\label{sec:level5}Application in Artificial Neural Network Regression}

Each of the aforementioned \textbf{SFC-M}  representations was used as an input to train three artificial neural networks (ANNs). The ANNs used in this research were multi-layer perceptrons (MLPs) as implemented by Pyzer-Knapp \textit{et al.}\cite{pyzerknapp_learning_2015,maml}.

MLPs perform supervised learning tasks and are known for being highly predictive. MLPs consist of three types of neuron layers: the input layer, which reads in feature representations; the multiple hidden layers, which process inputs with the use of nonlinear activation functions; and the output layer, which combines inputs into results. At each neuron, inputs are combined with weights, passed through an activation function, and connected with further neuron.

The activation function used in these MLPs is the following non-linear logistic function:
\begin{equation}
    \varphi(\alpha) = \frac{1}{1 + e^{-\alpha}}
\end{equation}

MLPs have several hyperparameters, which can be varied to optimize predictive ability. The most significant hyperparameters are:

\begin{itemize}
    \item architecture: the number of hidden layers in the network and the number of neurons in each layer,
    \item learn rate: the rate at which the weights, which are combined with inputs at each neuron, are updated,
    \item batch size: the size of the batch of the train set that is sent through the network. Sending the entire train set through the network takes a long time and creates redundancies. Sending through smaller batches accelerates training and creates random noise, smoothing out the distribution and preventing overfitting.
\end{itemize}

The Hyperopt package\cite{hyperopt} was used to optimize relevant hyperparameters for each MLP. Hyperopt uses sequential model-based optimization (SMBO, also called Bayesian optimization), which is particularly useful for minimizing functions that are costly to evaluate (\textit{i.e.} the training of a neural network). SMBO allows users to specify the configuration space of hyperparameter values to search as a prior distribution. SMBO iterates between fitting the model and then using that information to decide which hyperparameters to test next. The optimized hyperparameters for each MLP are collected in Table \ref{table:hyperparams}.

\begin{table}[h]
\centering
\caption[Optimized hyperparameters]{The optimized hyperparameters for each MLP.}
\begin{tabular}{ c c c c c }
\hline
\textbf{\textsf{Representation}} & \textbf{\textsf{$k$-Value}} & \textbf{\textsf{Architecture}}  & \textbf{\textsf{Learn Rate}} & \textbf{\textsf{Batch Size}} \\
\hline
\textsf{SFC-M1} & \textsf{7} & \textsf{[72, 38, 468]} & \textsf{3.74e-4} & \textsf{37} \\
\textsf{SFC-M2} & \textsf{10} & \textsf{[41, 62, 15]} & \textsf{7.11e-4} & \textsf{52} \\
\textsf{SFC-M3} & \textsf{17} & \textsf{[79, 33, 90]} & \textsf{1.03e-5} & \textsf{72} \\
\hline
\end{tabular}
\label{table:hyperparams}
\end{table}

Once hyperparameters were optimized, the MLPs were trained on 80\% of the dataset. Both \textbf{SFC-M} representations and energy targets were normalized. Furthermore, each MLP was trained simultaneously on all energy targets. Multi-task training prevents over-fitting of the model, allowing for a greater predictability on the test set\cite{multi_task}.

\section{\label{sec:level6}Results and Discussion}

Multi-layer perceptrons (MLPs) were used to predict calculated energies of organic molecular crystal structures using the three different \textbf{SFC-M} representations as inputs.

The predictive ability of the representations was quantitatively assessed using the mean absolute fraction error. The absolute fraction error, $E_{AF}$, considers the prediction error relative to the calculated value:
\begin{equation}
    E_{AF} = \Bigg \langle\left|{\frac{Y_{pred} - Y_{calc}}{Y_{calc}}}\right|\Bigg \rangle
    \label{eq:afe}
\end{equation}
where $Y_{calc}$ are the calculated values and $Y_{pred}$ are the predicted values. This metric is unitless.

As no previous studies have used machine learning  to predict electronic properties of organic molecular crystal structures, the mean predictor, which always predicts the average value of the training set, is used as a simple baseline. The absolute fraction errors for all representations are collected in Table \ref{table:maes}. All errors are produced using 10-fold cross validation.

\begin{table}[h]
\centering
\caption[Mean absolute errors]{The absolute fraction error for all \textbf{SFC-M} representations, as defined in Equation \ref{eq:afe} when predicting Single Point, Ewald, Lattice, and MBD energies.  These values are unitless}
\begin{tabular}{ c c c c c }
\hline
\hline
\multicolumn{5}{ c }{\textbf{\textsf{Mean Predictor Error }}} \\
\hline
\textbf{\textsf{FP}} & \textbf{\textsf{Single Point}} & \textbf{\textsf{Ewald}} & \textbf{\textsf{Lattice}} & \textbf{\textsf{MBD}} \\
\hline
- & \textsf{0.40} $\pm$ \textsf{0.03} & \textsf{0.59} $\pm$ \textsf{0.03} & \textsf{0.28} $\pm$ \textsf{0.04} & \textsf{0.88} $\pm$ \textsf{0.08} \\
\hline
\hline
\multicolumn{5}{ c }{\textbf{\textsf{Train Error}}} \\
\hline
\textbf{\textsf{FP}} & \textbf{\textsf{Single Point}} & \textbf{\textsf{Ewald}} & \textbf{\textsf{Lattice}} & \textbf{\textsf{MBD}} \\
\hline
\textsf{SFC-M1} & \textsf{0.28} $\pm$ \textsf{0.05} & \textsf{0.45} $\pm$ \textsf{0.10} & \textsf{0.25} $\pm$ \textsf{0.05} & \textsf{0.52} $\pm$ \textsf{0.09} \\
\textsf{SFC-M2} & \textsf{0.24} $\pm$ \textsf{0.03} & \textsf{0.34} $\pm$ \textsf{0.04} & \textsf{0.23} $\pm$ \textsf{0.02} & \textsf{0.49} $\pm$ \textsf{0.07} \\
\textsf{SFC-M3} & \textsf{0.14} $\pm$ \textsf{0.004} & \textsf{0.36} $\pm$ \textsf{0.01} & \textsf{0.23} $\pm$ \textsf{0.01} & \textsf{0.53} $\pm$ \textsf{0.02} \\
\hline
\hline
\multicolumn{5}{ c }{\textbf{\textsf{Test Error }}} \\
\hline
\textbf{\textsf{FP}} & \textbf{\textsf{Single Point}} & \textbf{\textsf{Ewald}} & \textbf{\textsf{Lattice}} & \textbf{\textsf{MBD}} \\
\hline
\textsf{SFC-M1} & \textsf{0.30} $\pm$ \textsf{0.03} & \textsf{0.54} $\pm$ \textsf{0.12} & \textsf{0.27} $\pm$ \textsf{0.07} & \textbf{\textsf{0.55}} $\pm$ \textsf{0.07} \\
\textsf{SFC-M2} & \textsf{0.27} $\pm$ \textsf{0.04} & \textsf{0.38} $\pm$ \textsf{0.06} & \textsf{0.24} $\pm$ \textsf{0.03} & \textsf{0.56} $\pm$ \textsf{0.07} \\
\textsf{SFC-M3} & \textbf{\textsf{0.16}} $\pm$ \textsf{0.01} & \textbf{\textsf{0.37}} $\pm$ \textsf{0.03} & \textbf{\textsf{0.22}} $\pm$ \textsf{0.02} & \textsf{0.58} $\pm$ \textsf{0.05} \\
\hline
\hline
\end{tabular}
\label{table:maes}
\end{table}

The absolute fraction error for all \textbf{SFC-M} representations are significantly lower than that of the mean predictor. These results indicate that the MLPs are producing meaningful predictions. Additionally, the train and test errors are comparable, suggesting that the MLPs are not overfitting.

As illustrated in Table \ref{table:maes}, \textbf{SFC-M3} is most predictive of the single point, Ewald summed, and lattice energies because its atomic descriptor contains information about the Coulombic charge of atoms. However, its atomic descriptor does not include information about the nuclear charge of the atoms. Therefore, it cannot be as predictive of the MBD energy. \textbf{SFC-M2} is generally more predictive than \textbf{SFC-M1} because its atomic descriptor contains information about the coordination sites of atoms in addition to their nuclear charge. The MBD energy is hard to predict using this method, but we are confident that with further investigation into the atomic descriptor dimension, progress could be achieved.

\begin{figure}[h]
\includegraphics[width=\columnwidth]{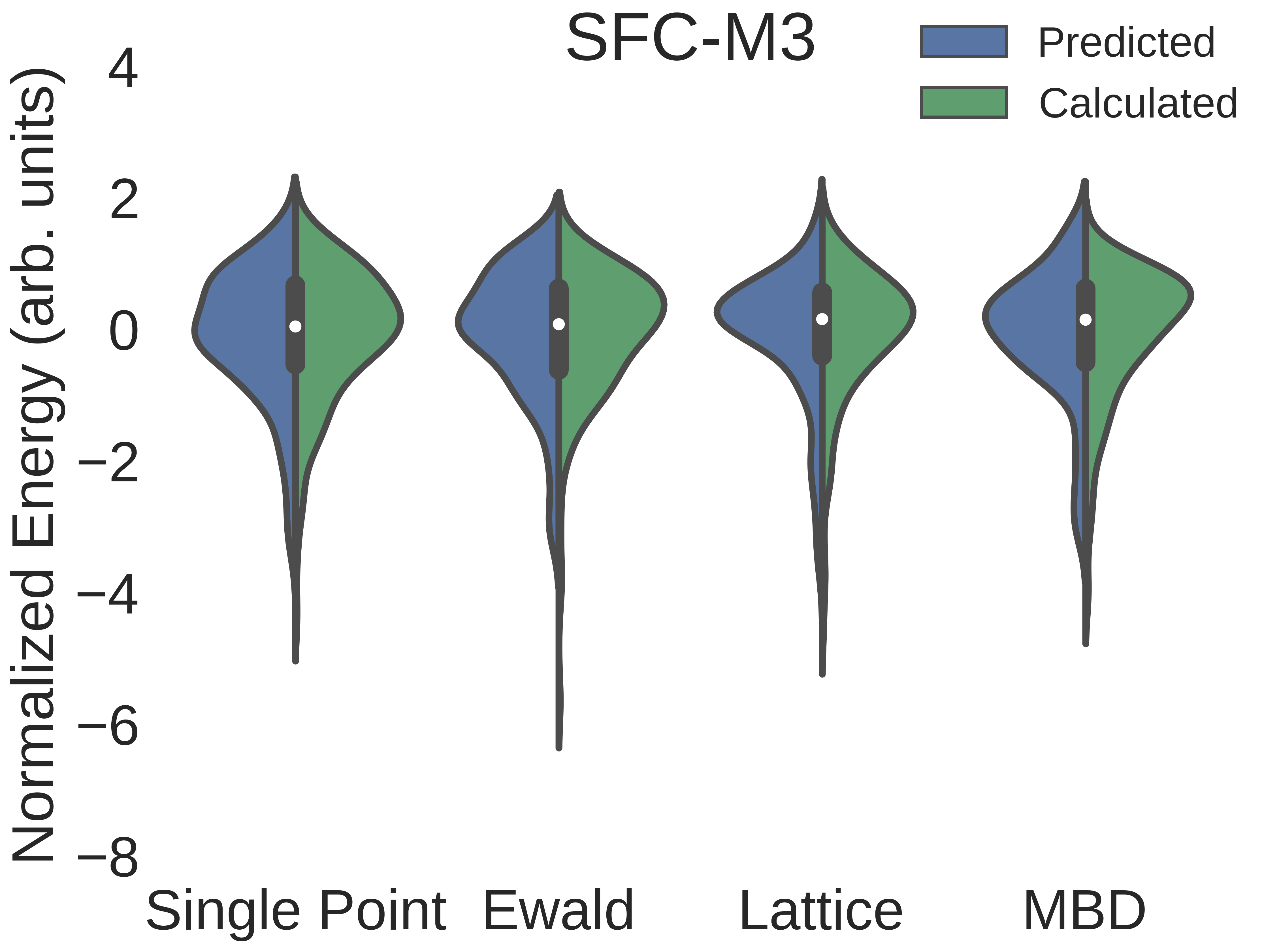}
\caption[Violin Plot]{The normalized distributions for single point, Ewald, lattice, and many-body dispersion energies of the unit cells from the test set. Distributions in green are calculated using DFT while distributions in blue are predicted using \textbf{SFC-M3}.  These values are unitless.
\label{fig:violin}}
\end{figure}

Fig. \ref{fig:violin} illustrates the distributions of the calculated values and the predicted values by \textbf{SFC-M3}.
The green, calculated distributions correspond to the blue, predicted distributions, especially for the single point and Ewald energies. As \textbf{SFC-M3} is the most predictive, this shows the potential of the \textbf{SFC-M} representations.

In the `computational funnel' paradigm often used in high-throughput materials discovery\cite{EOPK_HTVS}, the ability to correctly identify the most promising candidates is tantamount.  We can quantify the success of a method to achieve this goal through its recall rate.  The \textit{Nth} percent recall rate is defined as the number of candidates predicted to be contained within the top \textit{N} percent of a set, which are also contained within the top \textit{N} percent of the set, when ranked using the ground truth.  Another area in which recall rate is also important is crystal structure prediction where a funneling procedure is used, typically from charge-based energy to multipole-based energy to DFT energy\cite{csp}.  

The 15 \% recall rates for single point energies of the test set are collected in Table \ref{table:recall}. As a benchmark, we include the recall values for the calculated lattice energy ranking as a predictor of the DFT total energy ranking as this is a common funnel employed within crystal structure prediction. All recall rates are produced using 10-fold cross validation.

\begin{table}[h]
\centering
\caption[Recall rates]{The 15\% recall rates for all \textbf{SFC-M} representations when recalling the top single point energies in the test set when performed over 10-fold validation.  We report the best performing recall as an indicator of potential, the mean performance over the 10 folds, and their standard deviation.The recall when lattice energy is used as an ersatz for DFT single-point energy (labelled Lattice) is included as a benchmark.}
\begin{tabular}{ c c c c}
\hline
\textbf{\textsf{Feature}} & \textbf{\textsf{Best}} & \textbf{Mean} & \textbf{Std Dev.}\\
\hline
\textsf{Lattice}& 0.50 & 0.42 & 0.08\\
\textsf{SFC-M1}& 0.62 & 0.51 & \textbf{0.06}\\
\textsf{SFC-M2}& 0.69 & 0.54 & 0.08\\
\textsf{SFC-M3}& \textbf{0.88}& \textbf{0.68} & 0.08\\
\hline
\end{tabular}
\label{table:recall}
\end{table}

It can be seen that the recall rate for \textbf{SFC-M3} is significantly better than both \textbf{SFC-M1,2} and the multipole-based lattice energy for ranking the stability of molecular crystal structures, illustrating its utility for screening crystal structures as part of a computational funnelling scheme. 

\section{Conclusion}

This study explores a novel family of feature representation, \textbf{SFC-M}, for organic molecular crystal structures. This feature representation is used with multi-layer perceptrons to predict single-point, Ewald summed, lattice, and many-body dispersion energies.

This study shows that the \textbf{SFC-M} representations can be used in machine learning methods to make rapid predictions about electronic properties for organic molecular solids. These predictions are two or three orders of magnitude faster than calculations from traditional quantum mechanical methods, and they are relatively accurate. Especially for high-throughput endeavors, like the CEP, these machine learning models allow for the evaluation of solid-state properties of a large number of candidates. They can be used as an early step because of their small computational cost. Computationally intensive methods that provide more accurate properties can be used on candidates that are deemed promising by the predictions of the model.

Furthermore, the \textbf{SFC-M} representations are data-driven yet general, providing a new way to consider feature representations for crystal structures. As described in the methods, the \textbf{SFC-M} atomic descriptor dimension can be defined in a variety of ways and adjusted according to the property to be predicted. Moreover, the \textbf{SFC-M} can be generalized to map any $N$-dimensional dataset to a one-dimensional vector. Additional dimensions can be used to further describe atoms and their interactions with others in the crystal structure. Finally, this feature representation is independent from the unit cell size or type as long as a variety of sizes and types are well represented in the training set.

These results suggest that there is significant room to build on machine learning methods using crystal structure descriptors. The \textbf{SFC-M} representations can be expanded to include atoms beyond C, H, N, and O. It can also incorporate information about electron densities of molecules to allow for models that can predict electronic properties that depend on charge-charge and multipole interactions.

This study also explores Latent Semantic Indexing (LSI) in an unconventional chemical context and proposes a novel setting for the application of this method for the dimensionality reduction of sparse vectors. Much of the data used for machine learning is sparse due to the conversion of data from continuous to discrete. LSI allows for the compression of such data to a smaller number of real numbers, allowing it to be more useful for ML methods. Furthermore, the implementation of LSI implemented for this study \cite{maml} is built to scale only with the number of non-zero indices, rather than the more common scaling with total length of the sparse vector\cite{gensim}, thus allowing application to large data-sets.

The feature representations explored in this study can be used to train ML models that make rapid predictions about solid-state properties of molecules. This allows for the evaluation of these properties for a larger number of candidates than is possible using quantum mechanical methods, improving the robustness of high-throughput virtual screening projects such as the CEP and further accelerating the discovery of new materials in a variety of domains.

\section{Acknowledgements}

A.A.-G. and E.P.K. acknowledge support from The Department of Energy, Office of Basic Energy Sciences under award DE-SC0015959. A.A.-G. and D. R. acknoweldge support from the Department of Defense Vannevar Bush Fellowship under award number: N00014-16-1-2008. The authors are greatful for the exceptional support from Harvard FAS Research Computing and the use of the FAS Odyssey Cluster.

\bibliography{biblio}

\end{document}